\documentclass[conference]{IEEEtran}
\IEEEoverridecommandlockouts
\usepackage{cite}
\usepackage{amsmath,amssymb,amsfonts}
\usepackage{algorithmic}
\usepackage{graphicx}
\usepackage{textcomp}
\usepackage{xcolor}
\usepackage{booktabs}
\usepackage{multirow}
\usepackage{caption}
\usepackage{graphicx}
\usepackage{float}
\usepackage[utf8]{inputenc}
\usepackage[T1]{fontenc}
\usepackage{url}
\usepackage{tikz}
\usetikzlibrary{shapes.geometric, arrows.meta, positioning, calc}
\usepackage{adjustbox}

\tikzstyle{process} = [rectangle, rounded corners, minimum width=3cm, minimum height=1cm,text centered, draw=black, fill=blue!10]
\tikzstyle{labelbox} = [
    rectangle, 
    minimum width=1cm, 
    minimum height=0.8cm,
    text centered, 
    draw=black, 
    fill=green!15
]
\tikzstyle{outputbox} = [
    rectangle, 
    rounded corners, 
    minimum width=1cm, 
    minimum height=0.9cm, 
    text centered, 
    draw=black, 
    fill=orange!15, 
]
\tikzstyle{arrow} = [thick,->,>=Stealth]
\tikzstyle{smalltext} = [font=\footnotesize]
\def\BibTeX{{\rm B\kern-.05em{\sc i\kern-.025em b}\kern-.08em
    T\kern-.1667em\lower.7ex\hbox{E}\kern-.125emX}}
\begin{document}

\title{Divide, Cache, Conquer: Dichotomic Prompting\\for Efficient Multi-Label LLM-Based Classification\\
\thanks{Acknowledgement}
}


\author{
Mikołaj Langner, Jan Eliasz, Ewa Rudnicka, and Jan Kocoń \\
\textit{Department of Artificial Intelligence, Wroclaw Tech, Poland} \\
\texttt{\{mikolaj.langner, jan.eliasz, ewa.rudnicka, jan.kocon\}@pwr.edu.pl}
}

\maketitle

\begin{abstract}
We introduce a method for efficient multi-label text classification with large language models (LLMs), built on reformulating classification tasks as sequences of dichotomic (yes/no) decisions. Instead of generating all labels in a single structured response, each target dimension is queried independently, which, combined with a prefix caching mechanism, yields substantial efficiency gains for short-text inference without loss of accuracy. To demonstrate the approach, we focus on affective text analysis, covering 24 dimensions including emotions and sentiment. Using LLM-to-SLM distillation, a powerful annotator model (DeepSeek-V3) provides multiple annotations per text, which are aggregated to fine-tune smaller models (HerBERT-Large, CLARIN-1B, PLLuM-8B, Gemma3-1B). The fine-tuned models show significant improvements over zero-shot baselines, particularly on the dimensions seen during training. Our findings suggest that decomposing multi-label classification into dichotomic queries, combined with distillation and cache-aware inference, offers a scalable and effective framework for LLM-based classification. While we validate the method on affective states, the approach is general and applicable across domains.
\end{abstract}


\section{Introduction}
Affective computing, first articulated by Picard~\cite{picard1997affective}, has grown into a vibrant research field with applications in sentiment analysis, social media monitoring, and human–computer interaction. In the textual domain, annotating data for affective dimensions has proven useful across a wide range of tasks, yet it remains challenging due to the overlapping, context-dependent, and evolving nature of affective states~\cite{acheampong2020text,kmak2025predicting,ngo2025integrating, ferdinan2025fortifying, radlinski2025backtranslation, karlinska2024comprehensive, koptyra2024small, kocon2023deep, ngo2022studemo, milkowski2022multitask, kocon2019multilingual,kocon2019multi, gawron2021deep,korczynski2022compression,szolomicka2022multiaspectemo,kocon2019propagation,kocon2023differential, kocon2018context, kocon2021aspectemo}.  

A significant advancement would be the development of a universal classifier capable of detecting an arbitrary set of affective classes without retraining. Traditional classifiers with a fixed \textit{K}-class output are unsuitable for this purpose, as they offer neither flexibility in defining new classes nor customization of classification guidelines. Recent large language models (LLMs) have demonstrated strong zero-shot and few-shot capabilities~\cite{brown2020language,wei2022finetuned}, making them well-suited for such dynamic settings. However, directly deploying powerful LLMs can be prohibitively expensive, motivating the use of smaller language models (SLMs) trained via distillation.  

To address these challenges, we propose \textbf{dichotomic prompting}, which reformulates multi-label classification into independent yes/no queries and leverages prefix caching for efficiency. This strategy provides robustness in zero-shot settings while maintaining scalability.  

In summary, this paper makes the following contributions:
\begin{itemize}
    \item We introduce dichotomic prompting as an efficient mechanism for multi-label affective classification, yielding substantial inference gains for short texts when combined with prefix caching.
    \item We show that dichotomic prompting is particularly advantageous in zero-shot scenarios, mitigating errors from structured JSON generation and parsing.
    \item We empirically validate the approach on affective text analysis, demonstrating comparable or superior accuracy to structured prompting, with improved robustness and reduced computational cost.
\end{itemize}

\section{Background and Motivation}

\subsection{Limitations of Traditional Classification Architectures}

In affective computing, multi-label classification faces complexity due to overlapping, evolving, and context-sensitive labels. Traditional encoder-based models with fixed output heads (such as BERT~\cite{devlin2019bert}) are poorly suited for such settings. Extending their label space often demands retraining and architectural redesign, limiting adaptability and increasing computational burden.

This challenge has been long recognized in the broader multi-label learning literature. Early work formalized multi-label classification as a distinct problem setting~\cite{boutell2004mlc}, and surveys have emphasized the limitations of fixed-output models when label spaces evolve~\cite{tsoumakas2007overview,tsoumakas2009mining}. More recently, research on dynamic label taxonomies and label evolution has highlighted the same tension in modern architectures, where incorporating new or subjective categories often requires retraining from scratch~\cite{Yu2024PartialLabel,Tarekegn2024Survey}. Encoder–decoder models like T5~\cite{Raffel2020T5} introduce partial flexibility, but at a cost: their increased inference complexity presents practical limitations on deployment~\cite{Kementchedjhieva2023Exploration}.

\subsection{Decoder-Only LLMs for Prompt-Based Classification}

Decoder-only language models offer a more adaptable alternative. Label prediction can be formulated as answering natural-language queries rather than requiring fixed-size outputs. This makes the approach inherently flexible: new labels can be introduced simply by crafting new prompts, without retraining the model.

This idea builds on the success of prompt-based and instruction-tuned LLMs, which demonstrated strong zero-shot and few-shot capabilities~\cite{brown2020language,wei2022finetuned}. Systematic surveys of prompting methods~\cite{liu2023promptsurvey} further underscore how query-based reformulations enable robust generalization across subjective or sparsely represented dimensions. Consequently, decoder-only models are particularly well-suited to affective computing, where categories are often subjective and prone to overlap. Indeed, prior comparative reviews suggest that such models generalize more effectively than conventional supervised classifiers in low-resource and domain-adaptive settings~\cite{Galke2022Progress}.

\subsection{Prefix Caching for Efficient Prompt Execution}

Decoder-only transformer architectures commonly implement \textit{key–value (KV) caching} to optimize the computational efficiency of autoregressive generation. In this mechanism, the attention-related intermediate representations, specifically, the keys and values computed at each decoding step, are stored and reused in subsequent steps. This reuse obviates the need for recomputing these components during each forward pass, leading to substantial reductions in latency and compute overhead during inference~\cite{Sun2024YOCO}.

Building upon this foundation, inference-optimized frameworks such as \textbf{vLLM} introduce the concept of \textit{prefix caching}. This technique generalizes KV caching to enable reuse not only within a single decoding pass but across multiple prompts that share an identical input prefix~\cite{vLLMv1}. Prefix caching is particularly advantageous in settings characterized by high prompt redundancy, such as multi-label classification with prompt-based formulations. In such cases, distinct classification queries often share a large portion of their input prompt, differing only in task-specific suffixes (e.g., candidate labels or question variations), making them well-suited for cache reuse.

More broadly, the pursuit of efficient transformer inference has catalyzed a range of memory- and compute-aware optimizations. Notably, methods like FlashAttention~\cite{dao2022flashattention} re-engineer the attention mechanism itself to minimize memory access overhead and enhance throughput on modern hardware accelerators. These advancements are complementary to caching-based approaches, together contributing to a growing repertoire of scalable and latency-aware mechanisms for prompt execution.

In this work, we leverage these foundational techniques, particularly prefix caching, to support a novel dichotomic prompting formulation. This formulation entails executing a large number of structurally similar prompts efficiently, a task for which prefix caching offers substantial computational savings. Our approach situates itself within a broader trend of designing prompt strategies that are both algorithmically expressive and computationally tractable.

\section{Methodology}

\subsection{LLM-to-SLM Distillation}
\label{sec:distillation}

We adopt a knowledge distillation framework~\cite{Hinton2015}, wherein a large teacher model supervises smaller student models to reduce inference costs while preserving annotation quality. This paradigm is particularly effective in low-resource settings, where labeled data is scarce but powerful LLMs are available~\cite{Mansourian2025, Xu2024}.

Our teacher model, DeepSeek-V3~\cite{DeepSeek2024}, is a scalable mixture-of-experts decoder. It generates pseudo-labels for 24 affective dimensions using three independently sampled generations per input. Outputs are aggregated via majority vote to reduce variance and approximate a consensus annotation.

Pseudo-labels are produced using a fixed Polish-language prompt in structured JSON format:
\\
\begin{quote}
\small
\texttt{Your task is to identify emotions evoked by the given text.} \\
\texttt{1. For each emotion in the list: \{LABELS\}, assess whether the text evokes it (true/false).} \\
\texttt{2. Assess the overall tone of the text as one or more of: positive, negative, neutral.} \\
\texttt{The tone does not have to be exclusive—you may assign true to multiple options.} \\
\texttt{Return only a JSON object containing:} \\
\texttt{- all emotions with true/false values,} \\
\texttt{- three keys: positive, negative, neutral—also with true/false values.} \\
\texttt{Text: \{\{text\}\}}
\end{quote}
This prompt elicits structured JSON responses with binary judgments for each of the 24 affective dimensions, along with non-exclusive polarity labels. Its format ensures consistency and facilitates downstream parsing during label aggregation and model supervision.

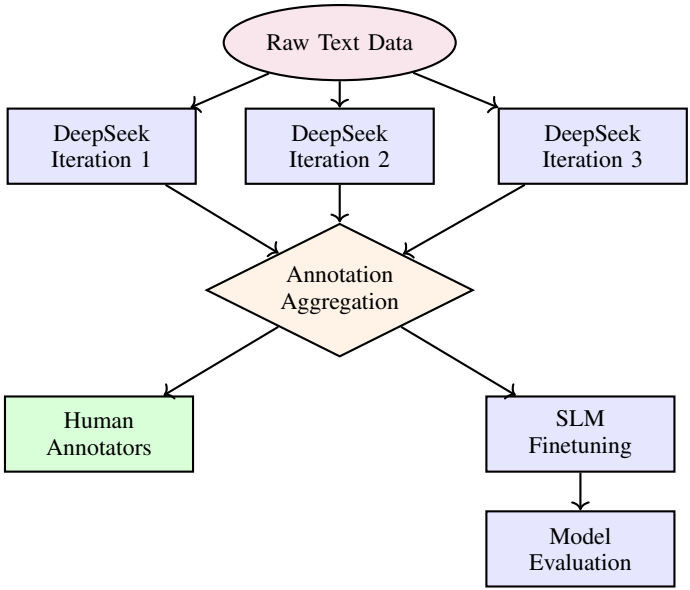
\begin{figure}[!tb]
\centering
\begin{tikzpicture}[
  node distance=0.5cm and 0.5cm,
  every node/.style={font=\small},
  process/.style={rectangle, draw=black, thick, minimum height=1.0cm, minimum width=2.5cm, align=center, fill=blue!10},
  cloud/.style={ellipse, draw=black, thick, minimum height=1cm, minimum width=3cm, align=center, fill=purple!10},
  aggregation/.style={diamond, draw=black, thick, aspect=2, align=center, minimum width=3cm, fill=orange!10},
  human/.style={rectangle, draw=black, thick, minimum height=1.0cm, minimum width=2.5cm, align=center, fill=green!15},
  arrow/.style={->, thick}
]

\node (data) [cloud] {Raw Text Data};

\node (deepseek1) [process, below left=of data, xshift=-0.3cm] {DeepSeek\\ Iteration 1};
\node (deepseek2) [process, below=of data, yshift=0.15cm] {DeepSeek\\ Iteration 2};
\node (deepseek3) [process, below right=of data, xshift=0.5cm] {DeepSeek\\ Iteration 3};

\node (aggregate) [aggregation, below=of deepseek2] {Annotation\\ Aggregation};

\node (human) [human, below=of aggregate, xshift=-3.2cm] {Human\\ Annotators};
\node (finetune) [process, below=of aggregate, xshift=3.2cm] {SLM\\ Finetuning};
\node (eval) [process, below=of finetune] {Model\\ Evaluation};

\draw[arrow] (data) -- (deepseek1);
\draw[arrow] (data) -- (deepseek2);
\draw[arrow] (data) -- (deepseek3);

\draw[arrow] (deepseek1) -- (aggregate);
\draw[arrow] (deepseek2) -- (aggregate);
\draw[arrow] (deepseek3) -- (aggregate);

\draw[arrow] (aggregate) -- (human);
\draw[arrow] (aggregate) -- (finetune);
\draw[arrow] (finetune) -- (eval);

\end{tikzpicture}
\caption{Annotation and distillation pipeline. Raw texts are independently annotated via three DeepSeek-V3 passes. The outputs are aggregated (via majority vote) to form consensus pseudo-labels. These annotations are then verified by human annotators for reliability assessment and used to fine-tune small language models (SLMs), which are subsequently evaluated.}

\label{fig:pipeline_overview}
\end{figure}

Figure~\ref{fig:pipeline_overview} illustrates the full annotation and distillation pipeline. Raw texts are first annotated by DeepSeek-V3 using the above prompt. The aggregated labels are then verified by human annotators for quality assessment and used to fine-tune four compact models: HerBERT-Large~\cite{HerBERT2021}, PLLuM-8B~\cite{PLLuM2024}, Gemma3-1B~\cite{Gemma32025}, and an internal CLARIN-1B model.

HerBERT-Large is an encoder-only Polish language model. PLLuM-8B and CLARIN-1B are decoder-only models adapted to Polish, while Gemma3-1B is a multilingual decoder-only model supporting over 140 languages, including Polish.

\subsection{Dichotomic Prompting Framework}

Traditional multi-label classification methods, such as fixed-output architectures or structured generation, predict all labels jointly but assume a static label space, limiting their use in dynamic tasks like affective computing.

We introduce \textit{dichotomic prompting}, which reformulates multi-label classification as $K$ independent binary decisions, each posed as a natural-language question, where the model responds with a single token (\texttt{Yes}/\texttt{No}) that maps directly to binary labels $(1, 0)$, respectively.

Key advantages include:
\begin{itemize}
    \item \textbf{Scalability}: Labels can be added or modified without retraining.
    \item \textbf{Flexibility}: Prompts can reflect subjective or evolving definitions.
    \item \textbf{Efficiency}: Only relevant dimensions need to be queried.
\end{itemize}

As shown in Figure~\ref{fig:prompting_flows}, dichotomic prompting differs from structured JSON prompting by issuing separate queries per label rather than generating all predictions in one pass.

Although this increases the number of forward passes, decoder-only models alleviate the cost via \textit{prefix caching}, which reuses shared prompt segments. While we demonstrate this method for affective classification in Polish, it generalizes across languages and domains.

\begin{figure}[!tb]
\centering

\begin{tikzpicture}[node distance=0.5cm and 0.5cm]

\node (title1) [smalltext, text width=7cm, align=center] {\textbf{Structured JSON Prompting}};
\node (input1) [process, below=of title1] {Prompt};
\node (prompt1) [labelbox, below=1cm of input1] {JSON Schema};
\node (output1) [outputbox, below=of prompt1, text width=5cm] {\{"Label $1$": true/false,\\ "Label $2$": true/false,\\ ...,\\ "Label $K$": true/false\}};

\node (title2) [smalltext, below=1.0cm of output1, text width=7cm, align=center] {\textbf{Dichotomic Prompting}};
\node (input2) [process, below=of title2] {Prompt};

\node (prompt2a) [labelbox, below=1cm of input2, xshift=-2cm] {Label $1$};
\node (prompt2b) [labelbox, below=1cm of input2, xshift=0cm] {Label $2$};
\node (prompt2c) [labelbox, below=1cm of input2, xshift=2cm] {Label $K$};

\node (output2a) [outputbox, below=of prompt2a] {Yes/No};
\node (output2b) [outputbox, below=of prompt2b] {Yes/No};
\node (output2c) [outputbox, below=of prompt2c] {Yes/No};

\draw [arrow] (input1) -- (prompt1);
\node[draw, fill=white, circle, inner sep=1pt, line width=0.3pt] at ($(input1)!0.5!(prompt1)$){\scriptsize\textbf{+}};
\draw [arrow] (prompt1) -- (output1);

\draw [arrow] (input2) -- (prompt2a);
\draw [arrow] (input2) -- (prompt2b);
\draw [arrow] (input2) -- (prompt2c);

\node[draw, fill=white, circle, inner sep=1pt, line width=0.3pt] at ($(input2)!0.5!(prompt2a)$) {\scriptsize\textbf{+}};
\node[draw, fill=white, circle, inner sep=1pt, line width=0.3pt] at ($(input2)!0.5!(prompt2b)$) {\scriptsize\textbf{+}};
\node[draw, fill=white, circle, inner sep=1pt, line width=0.3pt] at ($(input2)!0.5!(prompt2c)$) {\scriptsize\textbf{+}};
\node at ($(prompt2b)!0.5!(prompt2c)$) {\textbf{\dots}};

\draw [arrow] (prompt2a) -- (output2a);
\draw [arrow] (prompt2b) -- (output2b);
\draw [arrow] (prompt2c) -- (output2c);
\node at ($(output2b)!0.5!(output2c)$) {\textbf{\dots}};

\end{tikzpicture}
\caption{Prompting strategies for multi-label classification. \\ 
\textbf{Top:} Structured JSON prompting produces all label predictions in a single response by filling a predefined JSON schema. \\
\textbf{Bottom:} Dichotomic prompting issues one binary question per label and collects individual yes/no answers. Both strategies share the same input text but differ in how labels are queried and outputs are structured.}

\label{fig:prompting_flows}
\end{figure}

\subsection{Caching and Efficient Inference}

Although dichotomic prompting issues $K$ queries per input, one per label, the overhead is greatly reduced by \textit{prefix caching}, as supported in decoder-only models and optimized in inference frameworks such as vLLM~\cite{vLLMv1}. By reusing key–value attention states for shared prompt prefixes, redundant computation is avoided across similar queries.

In our setup, each prompt contains identical instructional and input text components, differing only in the target label. Decoder-only models process these prompts in two phases: a \textit{prefill phase} that encodes the shared prefix and a \textit{decode phase} that generates the final token. Since each response is constrained to a single-token output (\texttt{Yes}/\texttt{No}), nearly all computation lies in the prefill phase, making caching particularly effective.

\begin{figure}[!tb]
\begin{tikzpicture}[
  node distance=0.1cm,
  part/.style={rectangle, minimum height=1.2cm, minimum width=2.0cm, text centered, draw=black, thick},
  cached/.style={part, fill=green!15},
  uncached/.style={part, fill=gray!15},
  dimension/.style={part, fill=orange!15},
  label/.style={font=\small\bfseries}
]

\node[label] at (-3.3, 4.0) {Case 1};
\node[cached] (c1instr) at (-3.3, 3.0) {Instruction};
\node[dimension] (c1dim) [below=of c1instr] {Dimension};
\node[uncached] (c1text) [below=of c1dim] {Text};

\node[label] at (0, 4.0) {Case 2};
\node[cached] (c2text) at (0, 3.0) {Text};
\node[dimension] (c2dim) [below=of c2text] {Dimension};
\node[uncached] (c2instr) [below=of c2dim] {Instruction};

\node[label] at (3.3, 4.0) {Case 3};
\node[cached] (c3text) at (3.3, 3.0) {Text};
\node[cached] (c3instr) [below=of c3text] {Instruction};
\node[dimension] (c3dim) [below=of c3instr] {Dimension};

\end{tikzpicture}
\caption{Prompt structure configurations for evaluating prefix caching efficiency. Each column (Case 1–3) represents a different arrangement of prompt components: \textbf{Instruction}, \textbf{Text}, and \textbf{Target Label (Dimension)}. Green boxes denote cacheable segments reused across prompts; gray boxes indicate uncached, recomputed parts; and orange boxes highlight the position of the queried affective label. Case 3 maximizes cache utilization by placing both the instruction and input text in the shared prefix.}

\label{fig:caching_cases}
\end{figure}
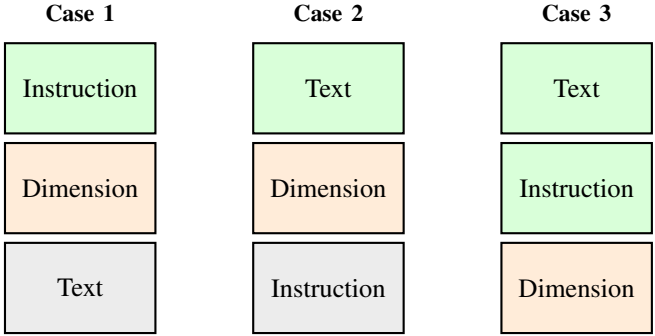

Prompt structure directly impacts cache efficiency. As illustrated in Figure~\ref{fig:caching_cases}, placing the label at the end of the prompt (Case 3) maximizes the cacheable region, as both instruction and text are reused across queries. This configuration was adopted in our main experiments.

\section{Experimental Setup}
\subsection{Dataset Composition}

To evaluate our framework, we compiled a 10{,}000-example dataset of affective Polish-language texts, drawn from six publicly available corpora. These include \texttt{wikinews\_pl}~\cite{wikinews2021}, \texttt{allegro\_reviews}~\cite{allegro2020}, \texttt{cdt}~\cite{cdt2013}, \texttt{open-coursebooks-pl}~\cite{coursebooks2021}, and a mixed-domain dataset~\cite{mms2023}, from which we extracted two subsets: \texttt{pl\_twitter\_sentiment} and \texttt{pl\_opi\_lil\_2012}. These sources span multiple domains and contexts of use.

Each text was annotated with binary labels across 24 affective dimensions. These dimensions are adapted from the 26-category framework introduced by \cite{mieleszczenko2023perspectives}, who systematically derived a fine-grained set of emotions and sentiment-related categories through large-scale questionnaires and annotation studies. This ensures that our schema remains grounded in established affective ontologies while remaining practical for large-scale Polish data.

Texts span five topical domains (\textit{politics}, \textit{sport}, \textit{science}, \textit{products}, \textit{culture}) and four usage contexts (\textit{news}, \textit{social media}, \textit{reviews}, \textit{coursebooks}). Table~\ref{tab:dataset_distribution} summarizes the distribution across sources, topics, and contexts.

The data is split into training, validation, and test sets with balanced label coverage. No preprocessing was applied to the text content.

\begin{table}[!tb]
\centering
\caption{Composition of the affective text dataset used in this study, grouped by source, topic, and context of use.}
\label{tab:dataset_distribution}
\begin{tabular}{llr}
\toprule
\textbf{Category} & \textbf{Type} & \textbf{Number of Texts} \\
\midrule
\multirow{6}{*}{Source Corpus}
    & \texttt{wikinews\_pl}             & 4,633 \\
    & \texttt{allegro\_reviews}         & 2,000 \\
    & \texttt{open-coursebooks-pl}     & 1,026 \\
    & \texttt{cdt}                      &   983 \\
    & \texttt{pl\_twitter\_sentiment}   &   713 \\
    & \texttt{pl\_opi\_lil\_2012}       &   645 \\
\midrule
\multirow{5}{*}{Topic}
    & Politics   & 2,000 \\
    & Sport      & 2,000 \\
    & Science    & 2,000 \\
    & Products   & 2,000 \\
    & Culture    & 2,000 \\
\midrule
\multirow{4}{*}{Usage Context}
    & News               & 4,633 \\
    & Social Media       & 2,341 \\
    & Review             & 2,000 \\
    & Coursebook         & 1,026 \\
\bottomrule
\end{tabular}
\end{table}

\subsection{Human and Model Annotation Agreement}

In addition to fully automated LLM-based experiments, we conducted a human verification study to assess the reliability of the obtained results. Two human annotators were employed to evaluate a subset of text fragments. Their task was to determine whether any of the predefined sentiment and emotion labels were applicable to each fragment.

To ensure consistency with the simple prompt provided to the model (Section~\ref{sec:distillation}), annotators were instructed to make intuitive judgments regarding the sentiment and emotional characteristics of the texts. No formal definitions or explanatory guidelines for the labels were provided. Instead, annotators were asked to identify all potential emotions or effects the text might evoke in a general reader.

In the first phase, annotators reviewed 40 text fragments accompanied by the binary label assignments produced by the LLM. They independently verified each label and corrected cases where their judgments diverged from the model’s outputs. The results were encouraging: both inter-annotator agreement and annotator-LLM agreement ranged between 70\% and 80\%.

In a subsequent experiment, we removed the model's annotations and asked the annotators to assign label values independently (blind condition). This procedure resulted in only a slight decrease in agreement levels but increased annotation time by approximately 50\%. Consequently, we adopted the initial procedure in which annotators directly verified and corrected the model’s outputs.

Following this protocol, the annotators proceeded to evaluate an additional 160 text fragments.

\subsection{Evaluation Protocol}
\label{sec:evaluation}

The fine-tuned models (HerBERT-Large, CLARIN-1B, PLLuM-8B, and Gemma3-1B) are trained using binary cross-entropy loss over 24 affective dimensions. Labels are derived from the LLM-driven distillation pipeline (Section~\ref{sec:distillation}) and serve as pseudo-labels generated for supervised training.

Model-specific hyperparameters are selected via 10-trial Bayesian optimization with early stopping on a held-out validation subset.

Evaluation is conducted under two regimes:
\begin{enumerate}
    \item \textbf{In-distribution (ID):} All 24 affective labels are available in the training, validation, and test sets.
    \item \textbf{Out-of-distribution (OOD):} A single label is excluded from the training and validation sets, and the model is evaluated only on that label in the test set. This leave-one-out setup is repeated across all labels.
\end{enumerate}

Gemma3-1B is used as the representative model for OOD evaluation, as it is the only small decoder-only model among those tested that reliably supports both prompting formats, including structured JSON outputs.

To assess the effects of fine-tuning (e.g., catastrophic forgetting or generalization), we compare zero-shot and fine-tuned OOD performance under both prompting strategies. Prompting instructions used during evaluation are paraphrased but format-consistent with those seen during training.

Evaluation metrics include macro- and micro-averaged F1 scores. Inference latency is also measured for Gemma3-1B under both prompting styles to estimate computational efficiency.

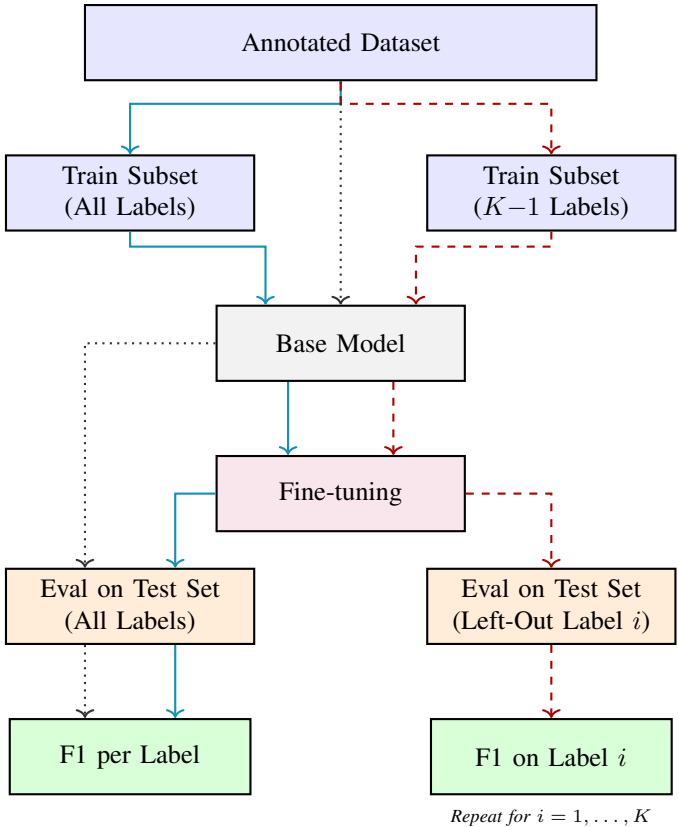
\begin{figure}[!tb]
\centering
\begin{tikzpicture}[
  node distance=0.5cm and 0.0cm,
  data/.style={rectangle, draw=black, thick, minimum width=3.3cm, minimum height=1.0cm, text centered, align=center, fill=blue!10},
  process/.style={rectangle, draw=black, thick, minimum width=3.3cm, minimum height=1.0cm, text centered, align=center, fill=gray!10},
  finetune/.style={rectangle, draw=black, thick, minimum width=3.3cm, minimum height=1.0cm, text centered, align=center, fill=purple!10},
  evaltype/.style={rectangle, draw=black, thick, minimum width=3.3cm, minimum height=1.0cm, text centered, align=center, fill=orange!15},
  result/.style={rectangle, draw=black, thick, minimum width=3.2cm, minimum height=1.0cm, text centered, align=center, fill=green!15},
  idarrow/.style={->, thick, color=cyan!70!black},
  oodarrow/.style={->, thick, dashed, color=red!70!black},
  zsarrow/.style={->, thick, dotted, color=black!80},
  note/.style={font=\scriptsize\itshape}
]

\node[data, minimum width=6.8cm] (dataset) at (0,7.0) {Annotated Dataset};

\node[data] (id_train) at (-2.8,5.0) {Train Subset\\ (All Labels)};
\node[data] (ood_train) at (2.8,5.0) {Train Subset\\ ($K{-}1$ Labels)};

\node[process] (base_model) at (0,3.0) {Base Model};

\node[finetune] (ft_block) at (0,1.0) {Fine-tuning};

\node[evaltype] (id_eval) at (-2.8,-0.5) {Eval on Test Set\\ (All Labels)};
\node[evaltype] (ood_eval) at (2.8,-0.5) {Eval on Test Set\\ (Left-Out Label $i$)};

\node[result] (id_f1) at (-2.8,-2.5) {F1 per Label};
\node[result] (ood_f1) at (2.8,-2.5) {F1 on Label $i$};

\draw[idarrow] (dataset.south) -- ++(0,-0.3) -| (id_train.north);
\draw[oodarrow] (dataset.south) -- ++(0,-0.3) -| (ood_train.north);
\draw[zsarrow] (dataset.south) -- ++(0,-0.3) -| (base_model.north);

\draw[idarrow] (id_train.south) -- ++(0,-0.2) -| ([xshift=-1.0cm]base_model.north);
\draw[oodarrow] (ood_train.south) -- ++(0,-0.2) -| ([xshift=1.0cm]base_model.north);

\draw[idarrow] ([xshift=-0.7cm]base_model.south) -- ++(0,-0.2) -| ([xshift=-0.7cm]ft_block.north);
\draw[oodarrow] ([xshift=0.7cm]base_model.south) -- ++(0,-0.2) -| ([xshift=0.7cm]ft_block.north);

\draw[idarrow] (ft_block.west) -| ([xshift=0.6cm]id_eval.north);
\draw[oodarrow] (ft_block.east)  -| (ood_eval.north);

\draw[idarrow] ([xshift=0.6cm]id_eval.south) -- ([xshift=0.6cm]id_f1.north);
\draw[oodarrow] (ood_eval) -- (ood_f1);

\draw[zsarrow] (base_model.west) -| ([xshift=-0.6cm]id_eval.north);
\draw[zsarrow] ([xshift=-0.6cm]id_eval.south) -| ([xshift=-0.6cm]id_f1.north);

\node[note] at (ood_f1.south) [yshift=-0.3cm, fill=white] {Repeat for $i = 1, \ldots, K$};
    
\end{tikzpicture}
\caption{Evaluation protocol for all settings.\\ \textbf{Solid cyan path:} Fine-tuning and evaluation on all labels. \\ \textbf{Dashed red path:} Leave-one-out (LOO) strategy where one label is excluded from training and evaluated separately. \\ \textbf{Dotted gray path:} Zero-shot evaluation using the base model without fine-tuning.}

\label{fig:eval_pipeline_final}
\end{figure}

Figure~\ref{fig:eval_pipeline_final} summarizes the evaluation setup across all configurations.

\section{Results}
\subsection{Annotation Agreement Analysis}

LLM-generated annotations exhibit strong alignment with human judgments. As shown in Table~\ref{tab:psa_agreement}, DeepSeek-V3 achieves high Positive Specific Agreement (PSA) scores against two human annotators. Inter-annotator agreement was similarly high, while intra-model consistency (inter-run) reached over $0.9$.

\begin{table}[!tb]
\centering
\caption{Positive Specific Agreement (PSA) scores comparing the consistency of affective annotations between human annotators and LLM-generated labels.}
\label{tab:psa_agreement}
\begin{tabular}{l c}
\toprule
\textbf{Comparison} & \textbf{PSA Score} \\
\midrule
LLM (inter-run average agreement) & 0.925 \\
Human Annotator A1 vs. A2 & 0.791 \\
LLM (aggregated) vs. Annotator A1 & 0.818 \\
LLM (aggregated) vs. Annotator A2 & 0.803 \\
\bottomrule
\end{tabular}
\end{table}

\begin{figure}[!tb]
    \centering
    \includegraphics[width=0.45\textwidth]{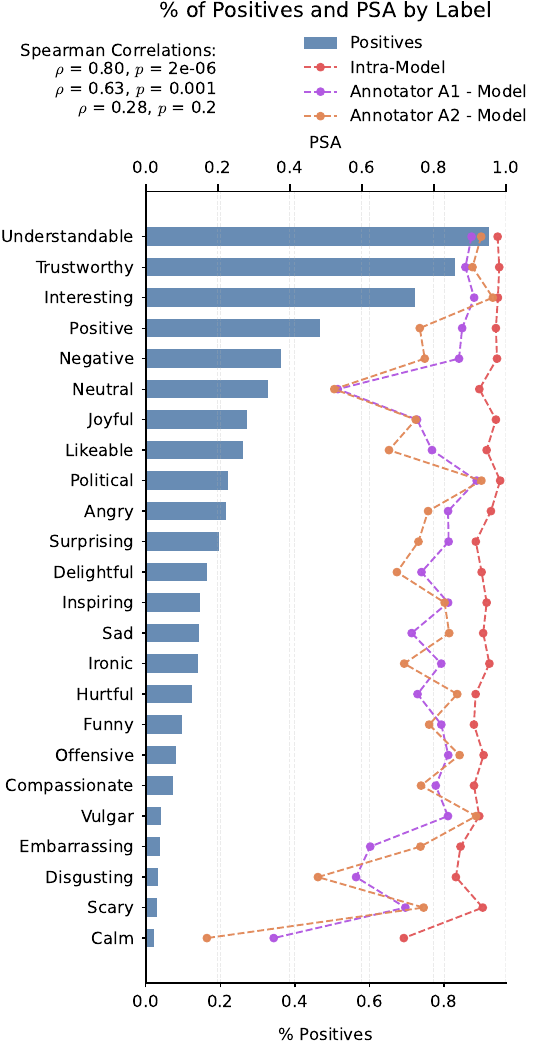}
    \caption{Relationship between label frequency and annotation agreement. Blue bars show the percentage of positive annotations per label, while dashed lines indicate Positive Specific Agreement (PSA): intra-model consistency (red), annotator A1 vs.\ model (purple), and annotator A2 vs.\ model (orange). Reported Spearman correlations (top left) quantify the relationship between label prevalence and PSA, showing that more frequent labels tend to yield higher agreement.}
    \label{fig:annotation}
\end{figure}

Figure~\ref{fig:annotation} shows the relationship between label prevalence and annotation agreement. We find a strong positive correlation between the frequency of positive labels and intra-model PSA ($\rho = 0.80$, $p < 0.001$), indicating that frequent categories are annotated more consistently. A similar, though weaker, correlation is observed between Annotator~A1 and the model ($\rho = 0.63$, $p = 0.001$), while no significant correlation is found for Annotator~A2.

\subsection{Classification Performance}

Table~\ref{tab:model_results} reports macro and micro F1 scores across all models and prompting strategies. Among fine-tuned models, decoder-only architectures perform on par with or slightly better than the encoder-only HerBERT. While PLLuM-8B achieves the highest scores, further analysis focuses on Gemma3-1B as a more computationally practical option.

CLARIN-1B, which lacks instruction tuning, could not reliably generate structured outputs and was evaluated only under dichotomic prompting.

Label-wise in-distribution results are shown in Table~\ref{tab:per_label_f1_grouped}, where dichotomic prompting performs comparably to JSON prompting across most affective dimensions and models, confirming its effectiveness despite its simplicity.

\begin{table}[!tb]
\centering
\caption{Per-label in-distribution F1 scores for each model and prompting configuration. Results are grouped by model and prompting setups, where applicable. CLARIN-1B could not support structured output generation.}
\label{tab:per_label_f1_grouped}
\begin{adjustbox}{max width=\linewidth}
\begingroup
\setlength{\tabcolsep}{3pt}
\begin{tabular}{lcccccc}
\toprule
 & \multicolumn{1}{c}{\textbf{HerBERT}} & \multicolumn{1}{c}{\textbf{CLARIN-1B}} & \multicolumn{2}{c}{\textbf{Gemma3-1B}} & \multicolumn{2}{c}{\textbf{PLLuM-8B}} \\
\cmidrule(lr){2-2}\cmidrule(lr){3-3}\cmidrule(lr){4-5}\cmidrule(lr){6-7}
 & \multicolumn{1}{c}{—} & \multicolumn{1}{c}{\textbf{Dichotomic}} & \multicolumn{1}{c}{\textbf{JSON}} & \multicolumn{1}{c}{\textbf{Dichotomic}} & \multicolumn{1}{c}{\textbf{JSON}} & \multicolumn{1}{c}{\textbf{Dichotomic}} \\
\midrule
Understandable & 0.963 & 0.965 & 0.960 & 0.959 & 0.965 & 0.964 \\
Trustworthy    & 0.962 & 0.958 & 0.929 & 0.971 & 0.967 & 0.968 \\
Interesting    & 0.942 & 0.938 & 0.930 & 0.953 & 0.948 & 0.951 \\
Positive       & 0.905 & 0.884 & 0.885 & 0.865 & 0.919 & 0.922 \\
Negative       & 0.881 & 0.853 & 0.809 & 0.813 & 0.911 & 0.903 \\
Neutral        & 0.810 & 0.774 & 0.733 & 0.783 & 0.821 & 0.848 \\
Joyful         & 0.860 & 0.830 & 0.812 & 0.828 & 0.881 & 0.898 \\
Likeable       & 0.832 & 0.834 & 0.797 & 0.812 & 0.851 & 0.867 \\
Political      & 0.902 & 0.934 & 0.979 & 0.970 & 0.940 & 0.953 \\
Angry          & 0.827 & 0.820 & 0.845 & 0.867 & 0.890 & 0.896 \\
Surprising     & 0.490 & 0.564 & 0.194 & 0.306 & 0.757 & 0.655 \\
Delightful     & 0.794 & 0.746 & 0.878 & 0.886 & 0.820 & 0.829 \\
Inspiring      & 0.802 & 0.773 & 0.607 & 0.657 & 0.830 & 0.841 \\
Sad            & 0.694 & 0.722 & 0.449 & 0.507 & 0.772 & 0.789 \\
Ironic         & 0.647 & 0.615 & 0.619 & 0.633 & 0.812 & 0.805 \\
Hurtful        & 0.759 & 0.736 & 0.764 & 0.770 & 0.821 & 0.798 \\
Funny          & 0.472 & 0.385 & 0.527 & 0.548 & 0.762 & 0.685 \\
Offensive      & 0.777 & 0.695 & 0.834 & 0.853 & 0.855 & 0.808 \\
Compassionate  & 0.685 & 0.676 & 0.662 & 0.689 & 0.822 & 0.829 \\
Vulgar         & 0.622 & 0.610 & 0.548 & 0.591 & 0.822 & 0.707 \\
Embarrassing   & 0.413 & 0.440 & 0.586 & 0.612 & 0.592 & 0.639 \\
Disgusting     & 0.469 & 0.585 & 0.626 & 0.635 & 0.666 & 0.529 \\
Scary          & 0.232 & 0.316 & 0.516 & 0.521 & 0.688 & 0.645 \\
Calm           & 0.000 & 0.070 & 0.200 & 0.208 & 0.343 & 0.427 \\
\bottomrule
\end{tabular}
\endgroup
\end{adjustbox}
\end{table}

\begin{table}[!tb]
\centering
\caption{In-distribution macro and micro F1 scores for all evaluated models under different prompting strategies. CLARIN-1B, lacking instruction tuning, were evaluated using dichotomic prompting only.}

\label{tab:model_results}
\begin{tabular}{llcc}
\toprule
\textbf{Model} & \textbf{Prompting Setup} & \textbf{F1 (Macro)} & \textbf{F1 (Micro)} \\
\midrule
\multirow{1}{*}{\textsc{CLARIN-1B}} 
    & Dichotomic & 0.718 & 0.856 \\
\midrule
\multirow{2}{*}{\textsc{PLLuM-8B}} 
    & JSON       & 0.812 & 0.897 \\
    & Dichotomic & 0.801 & 0.898 \\
\midrule
\multirow{2}{*}{\textsc{Gemma3-1B}} 
    & JSON       & 0.712 & 0.846 \\
    & Dichotomic & 0.722 & 0.866 \\
\midrule
\multirow{1}{*}{\textsc{HerBERT}} 
    & --- & 0.728 & 0.864 \\
\bottomrule
\end{tabular}
\end{table}

\begin{figure}[!tb]
    \centering
    \includegraphics[width=0.45\textwidth]{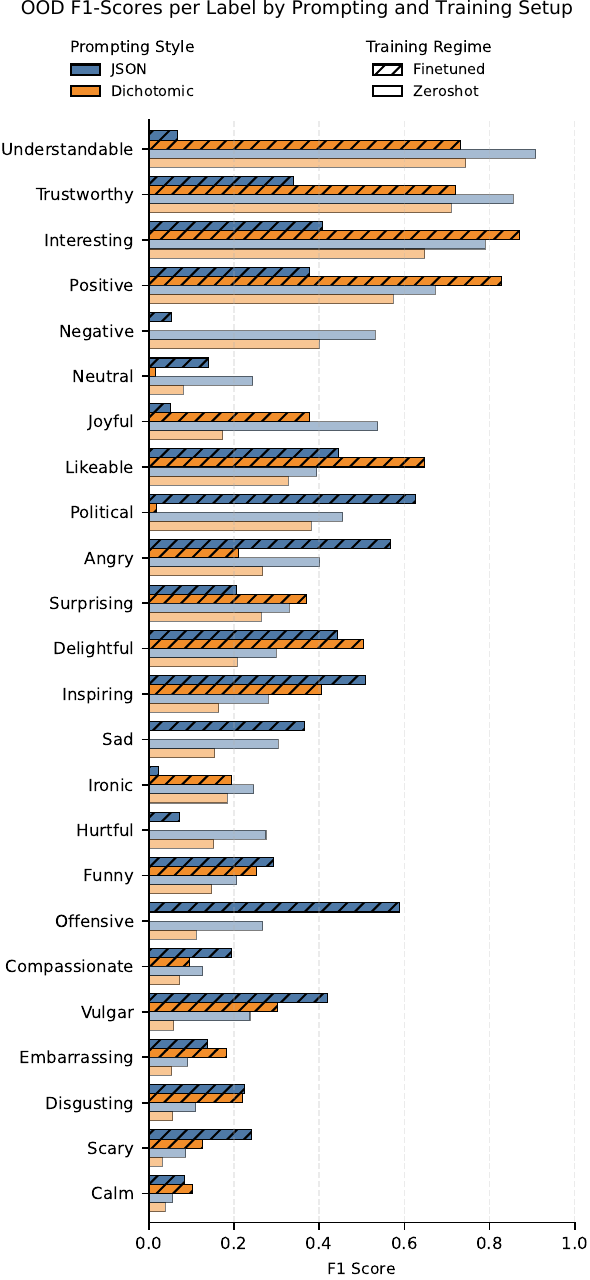}
    \caption{Per-label F1 scores on the Out-of-Distribution (OOD) setup using Gemma3-1B under different prompting styles and training regimes.}

    \label{fig:gemma}
\end{figure}

Figure~\ref{fig:gemma} presents per-label F1 scores for Gemma3-1B across prompting styles and training regimes. While fine-tuning improves many dimensions, it tends to reduce generalization to unseen labels. Notably, the variation in per-label outcomes further suggests that generalization gains are not strictly determined by label frequency.
In contrast, zero-shot prompting, particularly with dichotomic queries, achieves the strongest out-of-distribution performance compared to JSON prompting (Table~\ref{tab:ood_gemma_all}). 

These results indicate that dichotomic prompting yields more reliable zero-shot classification, overcoming common errors in structured output generation.




\begin{table}[!tb]
\centering
\caption{Out-of-distribution (OOD) classification performance of Gemma3-1B under different prompting and training regimes. F1 scores are reported for zero-shot and fine-tuned configurations, using both prompting strategies.}

\label{tab:ood_gemma_all}
\begin{tabular}{l l c c}
\toprule
\textbf{Training Regime} & \textbf{Prompting Setup} & \textbf{F1 (Macro)} & \textbf{F1 (Micro)} \\
\midrule
\multirow{2}{*}{Zero-shot}       & JSON  & 0.250 & 0.300 \\
& Dichotomic & 0.363 & 0.445 \\
\midrule
\multirow{2}{*}{Fine-tuned} & JSON  & 0.299 & 0.340 \\
 & Dichotomic & 0.286 & 0.178 \\
\bottomrule
\end{tabular}
\end{table}

Finally, we evaluated the impact of prompt structure on model performance using a Wilcoxon signed-rank test over F1 scores for different prompt formats illustrated in Figure~\ref{fig:caching_cases}. The ordering of prompt components had no significant effect on classification quality, confirming that dichotomic prompting is robust to structural variation.

\subsection{Inference Efficiency}
To compare the time efficiency of the two aforementioned prompting strategies, inference time was measured. The experiments were conducted on a system equipped with an NVIDIA L40 GPU featuring 45 GB of VRAM and under the CUDA version of 12.8. Inference was executed using the vLLM framework to ensure consistent model serving. Besides, vLLM supports \textit{automatic prefix caching}. A sample of 5,000 texts was used for evaluation, with text lengths ranging from 300 to 8,000 tokens. All the measurements were collected for the Gemma3-1B model.

\begin{figure}[!tb]
    \centering
    \includegraphics[width=0.5\textwidth]{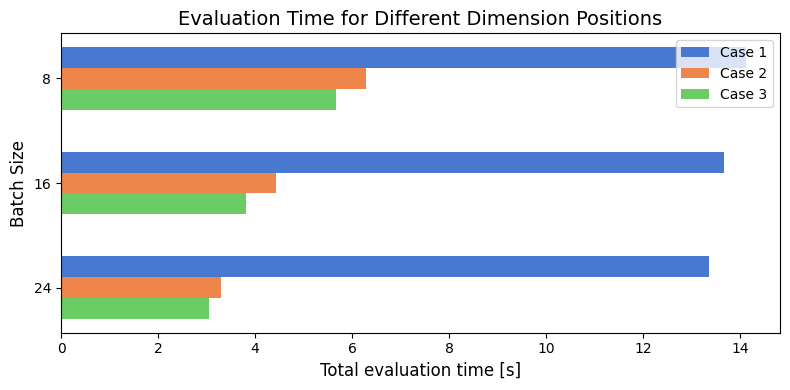}
    \caption{Comparison of evaluation times based on the placement of the affective dimension in the prompt using Gemma3-1B. Reported values correspond to the total evaluation time for a sample of 1,000 texts, each 300 tokens in length.}
    \label{fig:eff_case}
\end{figure}

The efficiency in the dichotomic prompting scenario varies significantly depending on the position of the affective dimension within the prompt, as presented in Figure~\ref{fig:eff_case}. This effect is primarily driven by the functioning of the prefix caching mechanism. The longer the shared prefix, the greater the portion of the prompt that can be cached, which in turn reduces the computational cost for subsequent prompts that share the same text but differ in the evaluated dimension.

The difference is particularly pronounced between scenarios where the text is not cached (Case 1) and those where it is (Case 2 and Case 3). Since the text typically constitutes the longest part of the prompt, placing the dimension before the text prevents caching of the majority of the input. Comparing Case 2 with Case 3 further confirms that positioning the dimension as close to the end of the prompt as possible yields faster evaluation, although the difference here is less substantial. This pattern remains consistent across different batch sizes. As our earlier experiments demonstrated that evaluation quality remains comparable across all three cases, we recommend adopting Case 3 as the preferred approach.

\begin{figure}[!tb]
    \centering
    \includegraphics[width=0.5\textwidth]{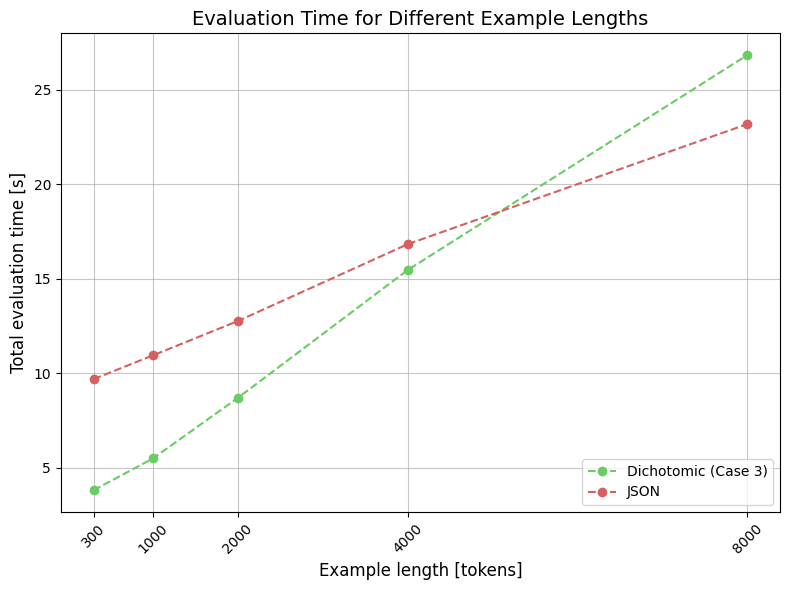}
    \caption{Comparison of evaluation times between Dichotomic and JSON prompting strategies using Gemma3-1B. Reported values correspond to the total evaluation time for a sample of 1,000 texts at each of the five examined text lengths.}

    \label{fig:eff_dicho_json}
\end{figure}

The comparative effectiveness of Dichotomic versus JSON prompting varies with the length of the input texts, as illustrated in Figure~\ref{fig:eff_dicho_json}. Dichotomic prompting is particularly efficient for shorter texts, but its advantage over JSON prompting diminishes as the input texts become longer. 


This suggests that Dichotomic prompting is advantageous for short texts, such as those typically used in affective evaluations, and in scenarios with limited computational resources, where batch sizes are necessarily small. We hypothesize that the diminishing advantage of Dichotomic prompting with longer texts and larger batch sizes is due to the limited capacity of the cache: longer texts cannot all be stored, and larger batches contain more texts, making it harder to keep them in the cache and thereby reducing the relative efficiency of Dichotomic prompting. A more detailed investigation of this phenomenon is left for future work.

\section{Discussion}

Our results highlight several key insights into dichotomic prompting for multi-label classification with decoder-only language models. Fine-tuned small models, especially PLLuM-8B, perform well on in-distribution data, but the generalization to out-of-distribution labels is limited. In contrast, zero-shot prompting (prior to fine-tuning) shows greater robustness to label exclusion. This illustrates the trade-off between in-domain optimization and generalization capacity.

Dichotomic prompting matches structured JSON prompting across models in both overall and per-label performance, despite its simpler format. This holds even under prompt structure variations, with no significant accuracy loss.

Efficiency-wise, dichotomic prompting benefits from prefix caching, enabling faster inference on short texts by reusing shared prompt segments across label queries. While structured prompting processes all labels in one pass, it lacks caching opportunities. For short inputs, dichotomic prompting is faster; however, its advantage wanes with longer texts. These trade-offs point to hybrid strategies, adapting the prompting format to input length or label set, as promising paths for deployment.

\section{Limitations}

Despite strong results, several limitations remain.

First, the annotation process depends entirely on a single teacher model, which, despite strong agreement with human annotations, may introduce biases or inconsistencies that propagate during training. Moreover, our human verification covered only a relatively small subset (200 texts), which may not be sufficient to fully capture potential systematic biases.

Second, the speedups from prefix caching are hardware- and system-dependent. Our setup favored dichotomic prompting for short inputs, but this advantage diminishes for longer texts due to cache capacity limits, and results may vary across platforms, models, or batch sizes.

Lastly, this work focuses on affective classification in Polish. While the method is language-agnostic in principle, its generalization to other domains, languages, or annotation schemas remains untested. The conclusions should thus be interpreted within this scope.

\section{Future Work}

Future work should explore scaling dichotomic prompting to tasks with larger label spaces. While this study focused on 24 affective dimensions, many applications, like topic classification or clinical coding, involve hundreds of labels. Understanding how performance and caching efficiency scale remains an open question.

Broader evaluation across languages and domains is also needed. Cross-lingual and cross-domain testing would provide insights into the method’s adaptability and robustness.

Hybrid prompting strategies also merit investigation. Dynamically choosing between structured and dichotomic setups based on input length, label count, or compute resources could optimize both speed and flexibility.

\section{Conclusion}

We presented a modular framework for multi-label classification using dichotomic prompting with decoder-only language models. By framing the task as a sequence of binary decisions and exploiting prefix caching, the approach balances accuracy, efficiency, and adaptability.

Experiments show that small models fine-tuned on LLM-generated labels can match or exceed zero-shot performance on in-domain data. Zero-shot prompting, meanwhile, generalizes better to unseen labels. Dichotomic prompting performs comparably to structured JSON prompting across all settings, while offering added benefits for short inputs.

Overall, the framework provides a scalable and efficient solution for multi-label classification, well-suited to dynamic label spaces and constrained compute environments.

\section*{Acknowledgments}
This work was financed by
(1) the European Regional Development Fund as part of the 2021–2027 European Funds for a Modern Economy (FENG) programme, project no. FENG.02.04-IP.040004/24: CLARIN – Common Language Resources and Technology Infrastructure;
(2) the European Regional Development Fund as a part of the 2014-2020 Smart Growth Operational Programme, project no. POIR.04.02.00-00C002/19: CLARIN – Common Language Resources and Technology Infrastructure;
(3) Digital Research Infrastructure for the Arts and Humanities DARIAH-PL (POIR.04.02.00-00-D006/20, KPOD.01.18-IW.03-0013/23);
(4) CLARIN ERIC – European Research Infrastructure Consortium: Common Language Resources and Technology Infrastructure (period: 2024-2026) funded by the Polish Minister of Science under the programme: "Support for the participation of Polish scientific teams in international research infrastructure projects", agreement number 2024/WK/01;
(5) Polish Minister of Digital Affairs under a special purpose subsidy No. 1/WII/DBI/2025: HIVE AI: Development and Pilot Deployment of Large Language Models in the Polish Public Administration;
(6) the statutory funds of the Department of Artificial Intelligence, Wroclaw University of Science and Technology.

AI-based tools, including ChatGPT, Grammarly Premium, and Writeful, were used exclusively to support linguistic clarity and improve the readability of the manuscript. 

\bibliographystyle{IEEEtran}
\bibliography{custom}

@book{picard1997affective,
  title={Affective computing},
  author={Picard, Rosalind W},
  year={2000},
  publisher={MIT press}
}

@article{acheampong2020text,
  title={Text-based emotion detection: Advances, challenges, and opportunities},
  author={Acheampong, Francisca Adoma and Wenyu, Chen and Nunoo-Mensah, Henry},
  journal={Engineering Reports},
  volume={2},
  number={7},
  pages={e12189},
  year={2020},
  publisher={Wiley Online Library}
}

@article{brown2020language,
  title={Language models are few-shot learners},
  author={Brown, Tom and Mann, Benjamin and Ryder, Nick and Subbiah, Melanie and Kaplan, Jared D and Dhariwal, Prafulla and Neelakantan, Arvind and Shyam, Pranav and Sastry, Girish and Askell, Amanda and others},
  journal={Advances in neural information processing systems},
  volume={33},
  pages={1877--1901},
  year={2020}
}

@inproceedings{
wei2022finetuned,
title={Finetuned Language Models are Zero-Shot Learners},
author={Jason Wei and Maarten Bosma and Vincent Zhao and Kelvin Guu and Adams Wei Yu and Brian Lester and Nan Du and Andrew M. Dai and Quoc V Le},
booktitle={International Conference on Learning Representations},
year={2022},
url={https://openreview.net/forum?id=gEZrGCozdqR}
}

@inproceedings{devlin2019bert,
  title={Bert: Pre-training of deep bidirectional transformers for language understanding},
  author={Devlin, Jacob and Chang, Ming-Wei and Lee, Kenton and Toutanova, Kristina},
  booktitle={Proceedings of the 2019 conference of the North American chapter of the association for computational linguistics: human language technologies, volume 1 (long and short papers)},
  pages={4171--4186},
  year={2019}
}

@article{boutell2004mlc,
  title={Learning multi-label scene classification},
  author={Boutell, Matthew R and Luo, Jiebo and Shen, Xipeng and Brown, Christopher M},
  journal={Pattern recognition},
  volume={37},
  number={9},
  pages={1757--1771},
  year={2004},
  publisher={Elsevier}
}

@article{tsoumakas2007overview,
  title={Multi-label classification: An overview},
  author={Tsoumakas, Grigorios and Katakis, Ioannis},
  journal={Data Warehousing and Mining: Concepts, Methodologies, Tools, and Applications},
  pages={64--74},
  year={2008},
  publisher={IGI Global}
}

@article{tsoumakas2009mining,
  title={Mining multi-label data},
  author={Tsoumakas, Grigorios and Katakis, Ioannis and Vlahavas, Ioannis},
  journal={Data mining and knowledge discovery handbook},
  pages={667--685},
  year={2010},
  publisher={Springer}
}

@article{Yu2024PartialLabel,
  title={Partial label learning with emerging new labels},
  author={Yu, Xiang-Ru and Wang, Deng-Bao and Zhang, Min-Ling},
  journal={Machine Learning},
  volume={113},
  number={4},
  pages={1549--1565},
  year={2024},
  publisher={Springer}
}

@article{Tarekegn2024Survey,
  title={Deep learning for multi-label learning: A comprehensive survey},
  author={Tarekegn, Adane Nega and Ullah, Mohib and Cheikh, Faouzi Alaya},
  journal={arXiv preprint arXiv:2401.16549},
  year={2024}
}

@article{Raffel2020T5,
  title={Exploring the limits of transfer learning with a unified text-to-text transformer},
  author={Raffel, Colin and Shazeer, Noam and Roberts, Adam and Lee, Katherine and Narang, Sharan and Matena, Michael and Zhou, Yanqi and Li, Wei and Liu, Peter J},
  journal={Journal of machine learning research},
  volume={21},
  number={140},
  pages={1--67},
  year={2020}
}

@inproceedings{Kementchedjhieva2023Exploration,
  title={An Exploration of Encoder-Decoder Approaches to Multi-Label Classification for Legal and Biomedical Text},
  author={Kementchedjhieva, Yova and Chalkidis, Ilias},
  booktitle={Findings of the Association for Computational Linguistics: ACL 2023},
  pages={5828--5843},
  year={2023}
}

@article{liu2023promptsurvey,
  title={Pre-train, prompt, and predict: A systematic survey of prompting methods in natural language processing},
  author={Liu, Pengfei and Yuan, Weizhe and Fu, Jinlan and Jiang, Zhengbao and Hayashi, Hiroaki and Neubig, Graham},
  journal={ACM computing surveys},
  volume={55},
  number={9},
  pages={1--35},
  year={2023},
  publisher={ACM New York, NY}
}

@article{Galke2022Progress,
  title={Are we really making much progress in text classification? A comparative review},
  author={Galke, Lukas and Diera, Andor and Lin, Bao Xin and Khera, Bhakti and Meuser, Tim and Singhal, Tushar and Karl, Fabian and Scherp, Ansgar},
  journal={arXiv preprint arXiv:2204.03954},
  year={2022}
}

@article{Sun2024YOCO,
  title={You only cache once: Decoder-decoder architectures for language models},
  author={Sun, Yutao and Dong, Li and Zhu, Yi and Huang, Shaohan and Wang, Wenhui and Ma, Shuming and Zhang, Quanlu and Wang, Jianyong and Wei, Furu},
  journal={Advances in Neural Information Processing Systems},
  volume={37},
  pages={7339--7361},
  year={2024}
}

@inproceedings{vLLMv1,
  title={Efficient memory management for large language model serving with pagedattention},
  author={Kwon, Woosuk and Li, Zhuohan and Zhuang, Siyuan and Sheng, Ying and Zheng, Lianmin and Yu, Cody Hao and Gonzalez, Joseph and Zhang, Hao and Stoica, Ion},
  booktitle={Proceedings of the 29th symposium on operating systems principles},
  pages={611--626},
  year={2023}
}

@article{dao2022flashattention,
  title={Flashattention: Fast and memory-efficient exact attention with io-awareness},
  author={Dao, Tri and Fu, Dan and Ermon, Stefano and Rudra, Atri and R{\'e}, Christopher},
  journal={Advances in neural information processing systems},
  volume={35},
  pages={16344--16359},
  year={2022}
}

@article{Hinton2015,
  title={Distilling the knowledge in a neural network},
  author={Hinton, Geoffrey and Vinyals, Oriol and Dean, Jeff},
  journal={arXiv preprint arXiv:1503.02531},
  year={2015}
}

@article{Mansourian2025,
  title={A Comprehensive Survey on Knowledge Distillation},
  author={Mansourian, Amir M and Ahmadi, Rozhan and Ghafouri, Masoud and Babaei, Amir Mohammad and Golezani, Elaheh Badali and Ghamchi, Zeynab Yasamani and Ramezanian, Vida and Taherian, Alireza and Dinashi, Kimia and Miri, Amirali and others},
  journal={arXiv preprint arXiv:2503.12067},
  year={2025}
}

@article{Xu2024,
  title={A survey on knowledge distillation of large language models},
  author={Xu, Xiaohan and Li, Ming and Tao, Chongyang and Shen, Tao and Cheng, Reynold and Li, Jinyang and Xu, Can and Tao, Dacheng and Zhou, Tianyi},
  journal={arXiv preprint arXiv:2402.13116},
  year={2024}
}

@article{DeepSeek2024,
  title={Deepseek-v3 technical report},
  author={Liu, Aixin and Feng, Bei and Xue, Bing and Wang, Bingxuan and Wu, Bochao and Lu, Chengda and Zhao, Chenggang and Deng, Chengqi and Zhang, Chenyu and Ruan, Chong and others},
  journal={arXiv preprint arXiv:2412.19437},
  year={2024}
}

@inproceedings{HerBERT2021,
  title={HerBERT: Efficiently Pretrained Transformer-based Language Model for Polish},
  author={Mroczkowski, Robert and Rybak, Piotr and Wr{\'o}blewska, Alina and Gawlik, Ireneusz},
  booktitle={Proceedings of the 8th Workshop on Balto-Slavic Natural Language Processing},
  pages={1--10},
  year={2021}
}

@misc{PLLuM2024,
  author       = {{PLLuM Consortium}},
  title        = {PLLuM: A Family of Polish Large Language Models},
  year         = {2025},
  howpublished = {\url{https://huggingface.co/CYFRAGOVPL}},
}

@inproceedings{kmak2025predicting,
  title={Predicting stock prices with ChatGPT-annotated Reddit sentiment: Hype or reality?},
  author={Kmak, Mateusz and Chmurzy{\'n}ski, Kamil and Matejuk, Kamil and Kotzbach, Pawe{\l} and Koco{\'n}, Jan},
  booktitle={International Conference on Computational Science},
  pages={307--322},
  year={2025},
  organization={Springer}
}

@article{ngo2025integrating,
  title={Integrating personalized and contextual information in fine-grained emotion recognition in text: A multi-source fusion approach with explainability},
  author={Ngo, Anh and Koco{\'n}, Jan},
  journal={Information Fusion},
  volume={118},
  pages={102966},
  year={2025},
  publisher={Elsevier}
}

@inproceedings{radlinski2025backtranslation,
  title={Backtranslation and Paraphrasing in the LLM Era? Comparing Data Augmentation Methods for Emotion Classification},
  author={Radli{\'n}ski, {\L}ukasz and Gu{\'s}ciora, Mateusz and Koco{\'n}, Jan},
  booktitle={International Conference on Computational Science},
  pages={3--17},
  year={2025},
  organization={Springer}
}

@inproceedings{karlinska2024comprehensive,
  title={Comprehensive Sentiment Analysis of Polish Book Reviews Using Large and Small Language Models},
  author={Karli{\'n}ska, Agnieszka and Mi{\l}kowski, Piotr and Czwordon-Lis, Paulina and Koptyra, Bart{\l}omiej and Koco{\'n}, Jan},
  booktitle={2024 IEEE International Conference on Data Mining Workshops (ICDMW)},
  pages={453--462},
  year={2024},
  organization={IEEE}
}

@inproceedings{gawron2021deep,
  title={Deep neural language-agnostic multi-task text classifier},
  author={Gawron, Karol and Pogoda, Micha{\l} and Ropiak, Norbert and Sw{\k{e}}drowski, Micha{\l} and Koco{\'n}, Jan},
  booktitle={2021 International Conference on Data Mining Workshops (ICDMW)},
  pages={136--142},
  year={2021},
  organization={IEEE}
}

@inproceedings{koptyra2024small,
  title={Small Language Models for Emotion Recognition in Polish Stock Market Investor Opinions},
  author={Koptyra, Bart{\l}omiej and Oleksy, Marcin and Dzi{\k{e}}cio{\l}, Ewa and Koco{\'n}, Jan},
  booktitle={2024 IEEE International Conference on Data Mining Workshops (ICDMW)},
  pages={463--470},
  year={2024},
  organization={IEEE}
}

@inproceedings{kocon2023deep,
  title={Deep emotions across languages: A novel approach for sentiment propagation in multilingual wordnets},
  author={Koco{\'n}, Jan},
  booktitle={2023 IEEE International Conference on Data Mining Workshops (ICDMW)},
  pages={744--749},
  year={2023},
  organization={IEEE}
}

@inproceedings{ngo2022studemo,
  title={Studemo: A non-aggregated review dataset for personalized emotion recognition},
  author={Ngo, Anh and Candri, Agri and Ferdinan, Teddy and Koco{\'n}, Jan and Korczynski, Wojciech},
  booktitle={Proceedings of the 1st Workshop on Perspectivist Approaches to NLP@ LREC2022},
  pages={46--55},
  year={2022}
}

@inproceedings{milkowski2022multitask,
  title={Multitask personalized recognition of emotions evoked by textual content},
  author={Mi{\l}kowski, Piotr and Saganowski, Stanis{\l}aw and Gruza, Marcin and Kazienko, Przemys{\l}aw and Piasecki, Maciej and Koco{\'n}, Jan},
  booktitle={2022 IEEE International Conference on Pervasive Computing and Communications Workshops and other Affiliated Events (PerCom Workshops)},
  pages={347--352},
  year={2022},
  organization={IEEE}
}

@inproceedings{kocon2019multilingual,
  title={Multilingual and language-agnostic recognition of emotions, valence and arousal in large-scale multi-domain text reviews},
  author={Koco{\'n}, Jan and Mi{\l}kowski, Piotr and Wierzba, Ma{\l}gorzata and Konat, Barbara and Klessa, Katarzyna and Janz, Arkadiusz and Riegel, Monika and Juszczyk, Konrad and Grimling, Damian and Marchewka, Artur and others},
  booktitle={Language and Technology Conference},
  pages={214--231},
  year={2019},
  organization={Springer}
}

@inproceedings{kocon2019multi,
  title={Multi-level analysis and recognition of the text sentiment on the example of consumer opinions},
  author={Koco{\'n}, Jan and Za{\'s}ko-Zieli{\'n}ska, Monika and Mi{\l}kowski, Piotr},
  booktitle={Proceedings of the International Conference on Recent Advances in Natural Language Processing (RANLP 2019)},
  pages={559--567},
  year={2019}
}

@article{ferdinan2025fortifying,
  title={Fortifying nlp models against poisoning attacks: The power of personalized prediction architectures},
  author={Ferdinan, Teddy and Koco{\'n}, Jan},
  journal={Information Fusion},
  volume={114},
  pages={102692},
  year={2025},
  publisher={Elsevier}
}

@article{Gemma32025,
  title={Gemma 3 technical report},
  author={Team, Gemma and Kamath, Aishwarya and Ferret, Johan and Pathak, Shreya and Vieillard, Nino and Merhej, Ramona and Perrin, Sarah and Matejovicova, Tatiana and Ram{\'e}, Alexandre and Rivi{\`e}re, Morgane and others},
  journal={arXiv preprint arXiv:2503.19786},
  year={2025}
}

@misc{wikinews2021,
  author       = {Rafa{\l} Po{\'s}wiata},
  title        = {wikinews-pl},
  year         = {2021},
  howpublished = {\url{https://huggingface.co/datasets/rafalposwiata/wikinews-pl}}
}

@inproceedings{allegro2020,
  title={KLEJ: Comprehensive Benchmark for Polish Language Understanding},
  author={Rybak, Piotr and Mroczkowski, Robert and Tracz, Janusz and Gawlik, Ireneusz},
  booktitle={Proceedings of the 58th Annual Meeting of the Association for Computational Linguistics},
  pages={1191--1201},
  year={2020}
}

@inproceedings{cdt2013,
  author       = {Michał Ptaszyński and Agata Pieciukiewicz and Paweł Dybała},
  title        = {Results of the PolEval 2019 Shared Task 6: First Dataset and Open Shared Task for Automatic Cyberbullying Detection in Polish Twitter},
  booktitle    = {Proceedings of the PolEval 2019 Workshop},
  year         = {2019},
  address      = {Warsaw, Poland},
  publisher    = {Institute of Computer Science, Polish Academy of Sciences},
}

@misc{coursebooks2021,
  author       = {Rafa{\l} Po{\'s}wiata},
  title        = {open-coursebooks-pl},
  year         = {2021},
  howpublished = {\url{https://huggingface.co/datasets/rafalposwiata/open-coursebooks-pl}}
}

@misc{mms2023,
  author       = {{Brand24 AI Lab}},
  title        = {mms: Polish online media sentiment corpus},
  year         = {2023},
  howpublished = {\url{https://huggingface.co/datasets/Brand24-AI/mms}}
}

@inproceedings{mieleszczenko2023perspectives,
  title={Capturing human perspectives in NLP: Questionnaires, annotations, and biases.},
  author={Mieleszczenko-Kowszewicz, Wiktoria and Kanclerz, Kamil and Bielaniewicz, Julita and Oleksy, Marcin and Gruza, Marcin and Wozniak, Stanislaw and Dzieciol, Ewa and Kazienko, Przemyslaw and Kocon, Jan},
  booktitle={NLPerspectives@ ECAI},
  year={2023}
}

@inproceedings{korczynski2022compression,
  title={Compression methods for transformers in multidomain sentiment analysis},
  author={Korczy{\'n}ski, Wojciech and Koco{\'n}, Jan},
  booktitle={2022 IEEE International Conference on Data Mining Workshops (ICDMW)},
  pages={419--426},
  year={2022},
  organization={IEEE}
}

@inproceedings{szolomicka2022multiaspectemo,
  title={Multiaspectemo: Multilingual and language-agnostic aspect-based sentiment analysis},
  author={Szo{\l}omicka, Joanna and Kocon, Jan},
  booktitle={2022 IEEE International Conference on Data Mining Workshops (ICDMW)},
  pages={443--450},
  year={2022},
  organization={IEEE}
}

@inproceedings{kocon2019propagation,
  title={Propagation of emotions, arousal and polarity in WordNet using Heterogeneous Structured Synset Embeddings},
  author={Koco{\'n}, Jan and Janz, Arkadiusz},
  booktitle={Proc. of the 10th Global Wordnet Conference},
  pages={336--341},
  year={2019}
}

@inproceedings{kocon2023differential,
  title={Differential dataset cartography: Explainable artificial intelligence in comparative personalized sentiment analysis},
  author={Koco{\'n}, Jan and Baran, Joanna and Kanclerz, Kamil and Kajstura, Micha{\l} and Kazienko, Przemys{\l}aw},
  booktitle={International Conference on Computational Science},
  pages={148--162},
  year={2023},
  organization={Springer}
}

@inproceedings{kocon2018context,
  title={Context-sensitive sentiment propagation in wordnet},
  author={Koco{\'n}, Jan and Janz, Arkadiusz and Piasecki, Maciej},
  booktitle={Proceedings of the 9th global wordnet conference},
  pages={329--334},
  year={2018}
}

@inproceedings{kocon2021aspectemo,
  title={Aspectemo: multi-domain corpus of consumer reviews for aspect-based sentiment analysis},
  author={Koco{\'n}, Jan and Radom, Jarema and Kaczmarz-Wawryk, Ewa and Wabnic, Kamil and Zaj{\k{a}}czkowska, Ada and Za{\'s}ko-Zieli{\'n}ska, Monika},
  booktitle={2021 International Conference on Data Mining Workshops (ICDMW)},
  pages={166--173},
  year={2021},
  organization={IEEE}
}

\end{document}